# The object detection model uses combined extraction with KNN and RF classification

**Florentina Tatrin Kurniati[1,2], Daniel HF Manongga[1], Irwan Sembiring[1], Sutarto Wijono[3], Roy Rudolf Huizen[4]**

[1]Faculty of Information Technology, Universitas Kristen Satya Wacana, Salatiga, Indonesia
[2]Faculty of Informatics and Computer, Institut Teknologi dan Bisnis STIKOM, Bali, Indonesia
[3]Faculty of Psychology, Universitas Kristen satya Wacana, Salatiga, Indonesia
[4]Department of Magister Information System, Institut Teknologi dan Bisnis STIKOM, Bali, Indonesia

**ABSTRACT**

Object detection plays an important role in various fields. Developing detection models for 2D objects that experience rotation and texture variations is a challenge. In this research, the initial stage of the proposed model integrates the gray-level co-occurrence matrix (GLCM) and local binary patterns (LBP) texture feature extraction to obtain feature vectors. The next stage is classifying features using k-nearest neighbors (KNN) and random forest (RF), as well as voting ensemble (VE). System testing used a dataset of 4,437 2D images, the results for KNN accuracy were 92.7% and F1-score 92.5%, while RF performance was lower. Although GLCM features improve performance on both algorithms, KNN is more consistent. The VE approach provides the best performance with an accuracy of 93.9% and an F1-score of 93.8%, this shows the effectiveness of the ensemble technique in increasing object detection accuracy. This study contributes to the field of object detection with a new approach combining GLCM and LBP as feature vectors as well as VE for classification.

*Corresponding Author:*

Florentina Tatrin Kurniati
Faculty of Information Technology, Universitas Kristen Satya Wacana
Salatiga, Indonesia
Email: 982022026@student.uksw.edu

## 1. INTRODUCTION

The 2D digital image objects can be analyzed to obtain information or detect them. This model has been implemented in various technological fields including security systems, medical diagnosis and autonomous vehicle technology [1]. Analysis of 2D objects through feature extraction and classification processes. The detection process is based on specific extracted features from 2D objects and is recognized based on feature patterns. In the analysis of 2D objects, difficulties often arise which are influenced by several factors, including rotation problems and texture variations in images which cause a decrease in accuracy in the process of detecting a 2D image object [2]. The different orientations and variations in texture were a big challenge. Different textures on the same object can produce very varied appearances in digital images due to lighting [3], [4]. Exploration of texture-based extraction methods is proposed for improved accuracy. Reliable extraction methods include gray-level co-occurrence matrix (GLCM) and local binary patterns (LBP). This method is a texture-based extraction method, for the GLCM method the extraction process uses pixel analysis, with an iterative calculation approach based on certain pixel pairs and certain intensities [5]. GLCM analysis can also use orientations or angles of 0°, 45°, 90°, and 135° [6]. GLCM features include contrast, correlation, energy, homogeneity, and entropy. This feature is to determine the characteristics of the extracted 2D image.

Contrast calculates intensity variations between neighboring pixels. Homogeneity calculates the diversity value of pixel intensity distribution. Entropy determines the degree of randomness of pixel intensity. Correlation measures a particular pixel with its neighboring pixels at a certain distance. Meanwhile, energy is used to determine the uniformity or regularity of size in a 2D object [7]. The LBP method for extracting objects with rotation variance, reduces changes influenced by object orientation, by normalizing the resulting binary pattern so that it is not affected by rotation, allowing consistent texture identification even when the object is rotated [8].

Integrating GLCM and LBP features can produce comprehensive feature vectors for object detection, especially objects experiencing rotation and displaying complex texture variations. The proposes to combine these two methods not only increases the robustness to texture variations and rotation but also provides a deeper understanding of the texture characteristics of objects in 2D digital images [9]. Model test using 4,437 2D objects, machine learning classification using k-nearest neighbors (KNN) and random forest (RF) [10]-[13]. The selection of KNN and RF algorithms is based on the characteristics and nature of the data. KNN is a non-parametric algorithm for handling data that does not have certain distribution assumptions. In contrast, RF has the advantage of being effective in overcoming overfitting, using an ensemble method that combines the results of many decision trees. In addition, it can handle categorical and numerical data and missing values [14], [15]. Algorithm selection must be based on understanding the data and object problems so that it can produce the best model.

Several studies have been carried out applying the GLCM method for object extraction. Like research conducted by Saifullah and Drezewski [5], this research uses 6 features, namely energy (En), contrast (Ct), entropy (Et), variance (V), correlation (Cr), and homogeneity (H), with classification using SVM. GLCM is also applied in the medical field for the classification of white blood cells, such as research conducted by Saikia and Devi [16] the testing process uses analysis of variance (ANOVA) and the zero component analysis (ZCA) test. This method was evaluated using the KNN classifier, using the blood cell count and detection (BCCD) dataset. For the classification, four categories of white blood cells (WBC) are used, namely lymphocytes, monocytes, neutrophils, and eosinophils.

The use of GLCM extraction for skin disease detection was carried out by Reddy *et al.* [17]. This study proposes a skin disease detection framework that uses segmentation and feature extraction to classify disease lesions. The segmentation method used is optimized region growing with (GWO), while texture features are extracted using GLCM and wide line detector (WLD). An autoencoder-based classification model is used to analyze feature representation.

Applying the KNN classification by Arumugaraja *et al.* [18] to analyze and monitor the gait of people with arthritis, trauma, and degenerative movement disorders, the KNN learning model produces 99.4% accuracy in detecting knee pain. Still in the field of health, research conducted by Sallam *et al.* [19] discusses the increasing prevalence of acute lymphoblastic leukemia, the classification models used include RF, support vector machine (SVM), KNN, and Naive Bayes (NB) show that this method is effective in improving classification accuracy.

Apart from the medical field, KNN is applied to classify batik images as done by Rangkuti *et al.* [20] This research focuses on supporting batik image classification that is invariant to image rotation and scale. This algorithm uses several windows such as sizes 6×6, 9×9, 12×12, and 15×15 or a combination of windows. To recognize batik patterns automatically, this research applies a batik classification method using kNN and artificial neural network (ANN).

In the agricultural sector, the application of KNN classification such as research conducted by Hossain *et al.* [11] can be used to detect and classify citrus diseases and assess fruit quality. Apart from that, research conducted by Saleem *et al.* [12] evaluated the visual features of artificial leaves. Apart from the extraction method, the classification process using machine learning also affects the accuracy value. Such as research conducted by Han *et al.* [21] on the new decision-tree multi-class support vector machines (ML-DSVM) + algorithm which integrates neural networks and SVM+ classification in one framework.

Other researchers use SVM for classification, such as Aamir *et al.* [22] in medical research for brain tumor classification. The same research uses SVM for classification, such as research conducted by Wu *et al.* [15] on early identification of gray mold disease on strawberry leaves. Subsequent research used SVM classification as carried out by Shetty and Patil [23]who studied the oral cancer detection framework. The model is used to detect the data pre-processing stage, which is used to reduce noise and remove unwanted artifacts. The next stage uses segmentation to separate the background area of the object. In this study, SVM with CNN was used to make the final decision regarding the presence or absence of oral cancer.

In research conducted by Bao *et al.* [24] to identify necrotic areas in magnetic resonance imaging (MRI) images of chronic spinal injuries. This method focuses on the accurate and automatic location of necrotic areas on MRI images. The model was built using the GLCM extraction method, Gabor texture features, local binary pattern features, and superpixel areas. Next, use machine learning techniques, such as SVM and RF, to detect areas experiencing necrosis.

Studies on object detection focus on objects with a wide range of rotation and texture variations. To improve accuracy, this research combines feature vectors from GLCM and LBP with ensemble classification. The primary goal is to address variations in detected objects and the impact of feature vectors resulting from GLCM and LBP extractions. KNN and RF classification methods, along with voting ensemble (VE), are employed. The evaluation of the proposed model measures accuracy, precision, recall, and F1-score. The next section of this paper presents a review of related research literature and research methodology at the beginning. Subsequently, an analysis is conducted by comparing methods to determine the performance of the proposed model.

In research conducted by Sallam *et al.* [25] for diagnosis by applying machine learning for classification in detecting blood cell cancer using KNN, SVM, NB, and RF, the results of accuracy, sensitivity, and specificity respectively being 99.69 %, 99.5%, and 99%, after using the gray wolf optimization algorithm feature selection. In research conducted by Han *et al.* [21] regarding selecting the histological characteristics of blood cells using the enhanced gray wolf optimization (EGWO) algorithm, it was stated that selecting characteristics based on certain criteria as the best cluster center used the k-means clustering algorithm. Also including using RF, SVM, and NB, the results show that the proposed methodology achieves a high level of accuracy with a value of 99.22%, precision of 99%, and sensitivity of 99%.

## 2. RESEARCH METHOD

The 2D object detection model uses a feature extraction and classification approach. Extraction using GLCM and LBP while classification using KNN, RF, and VE. This model increases accuracy and robustness for rotation problems and texture variations in 2D objects [26]. The extraction method used is shown in Table 1.

Table 1. Feature extraction method

| No | Feature |
|---|---|
| 1 | GLCM |
| 2 | LBP |
| 3 | Combination of LBP and GLCM |

Testing the 2D object detection model uses the Figure 1 model, with extraction method variants in Table 1. Testing uses 4,437 datasets [27] divided by 90% as training data and 10% as test data. The proposed model is shown in Figure 1.

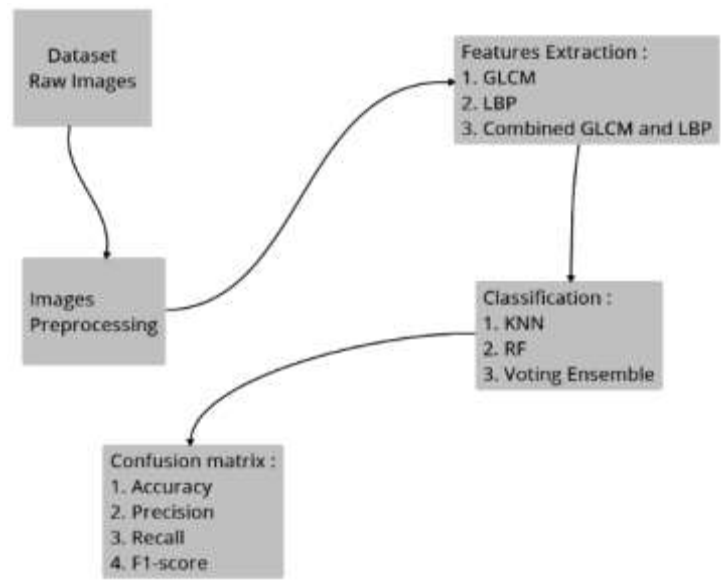

Figure 1. GLCM feature extraction model for 2D object classification

The proposed model for 2D object detection is shown in Figure 1. Each stage of the model is explained as follows; (i) pre-processing, at this stage there are 2 processes, namely resize and greyscale, resize is to adjust the size of the data into a more suitable form so that it can be used, while greyscale changes color images to greyscale images. (ii) at the feature extraction stage, using GLCM, LBP, and a combination of both. For GLCM features use contrast, correlation, energy, homogeneity, and entropy. The characteristics of each feature are explained as follows; contrast is the overall intensity of the relationship between a pixel and its neighbors [28], [29]. Contrast is formulated with the following in (1):

$$Contrast = \sum_{i,j=0}^{N-1} P_{ij}(i-j)^2 \qquad (1)$$

where $P_{ij}$ is an element contained in GLCM.

The correlation feature calculates a measure of the closeness of the relationship between pixels in the entire image. The proximity relationship between pixels is formulated in (2):

$$Correlation \sum_{i,j=0}^{N-1} \frac{P_{ij}(i-\mu)(j-\mu)}{\sigma^2} \qquad (2)$$

where μ is the mean GLCM, while σ² is the variance of the GLCM.

Energy is the square of the elements in GLCM, with a value between 0 and 1. This value indicates the level of uniformity of an image, formulated in (3):

$$Energy = \sum_{i,j=0}^{N-1} (P_{ij})^2 \qquad (3)$$

meanwhile, for homogeneity, it calculates the level of density of the elements distributed in the GLCM, shown in (4):

$$Homogeneity \sum_{i,j=0}^{N-1} \frac{P_{ij}}{1+(i-j)^2} \qquad (4)$$

meanwhile, entropy measures the level of uniformity between pixels in an image and its randomness, to calculate entropy it is shown in (5).

$$Entropy = \sum_{i,j=0}^{N-1} -In(P_{ij})P_{ij} \qquad (5)$$

In general, this equation explains that $P_{ij}$ is a GLCM that has been normalized symmetrically. With the notation N being the total number of gray levels in the image, and μ being the GLCM average. Meanwhile, the LBP method will compare each pixel with its neighbors in a certain environment. Specifically, LBP compares the value of a central pixel with the values of neighboring pixels in a local area around the pixel. This process involves selecting a pixel as the center and comparing it with its eight neighbors. If the value of a neighboring pixel is greater than or equal to the central pixel then it is given a value of 1, otherwise if it is smaller it will be given a value of 0. This process produces an 8-bit binary pattern for each pixel. This pattern is then converted into a decimal value which becomes the LBP value of the central pixel. This decimal value is calculated using the in (6).

$$LBP_{P_0} = \sum_{i=0}^{7} s(P_i - P_0) \times 2^i) \qquad (6)$$

where s(x) is a function that returns 1 if x≥0 and 0 if x<0, Pi represents the value of the neighboring pixel, and P0 denotes the value of the center pixel. The subsequent step involves classification utilizing KNN and RF. Visually, the classification through KNN is depicted in Figure 1.

Classification with kNN in the initial stage prepares the features used for classification. In this method, it is necessary to determine the k value of the number of nearest neighbors that will be used for the classification process [30]. The value of k is usually an odd number. Between data, distances are calculated, on all data. Calculate the distance using the Euclidean method shown in (7).

$$D(p,q) = sqrt((p1-q1)^2 + (p2-q2)^2 + \ldots + (pN-qN)^2) \qquad (7)$$

Where D is the distance, where p and q are two points calculated in the feature space, and N is the number of features. The next stage determines the value of k nearest neighbors obtained from the distance calculation. For classification, a new data class is determined based on the majority class of the k nearest neighbors, classification using the kNN method is shown in Figure 2.

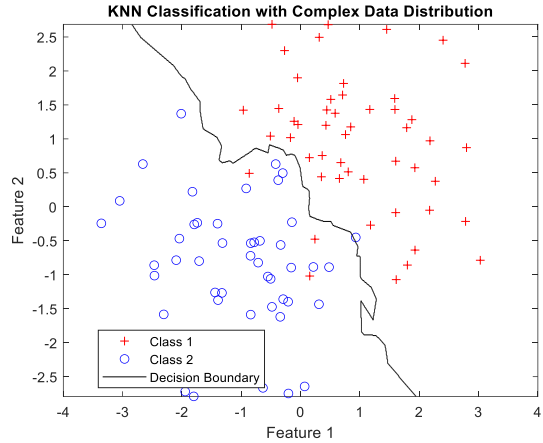

Figure 2. Classification with KNN

RF is a machine-learning algorithm, which can be used for classification, and regression. The algorithm combines many decision trees, during the training process. RF can also be implemented as ensemble machine learning, where several models are combined to improve performance. RF uses a sampling technique (bootstrapping), several samples are randomly selected from the training data, with replacement, to train each tree. Each bootstrap sample is used to train the decision tree. Splitting nodes to randomly select several features reduces the correlation between trees and increases diversity in the model, classification using the RF method is shown in Figure 3. In this model, classification is ensemble by combining KNN and RF, with a VE model.

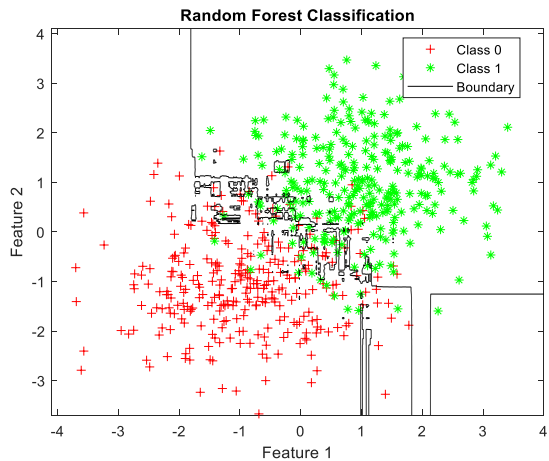

Figure 3. Classification with RF

## 3. RESULTS AND DISCUSSION

The test uses a 2D object dataset with each image rotated 5 degrees 9 times, with each having a diversity of textures, a total of 4,437 datasets divided into three classes shown in Figure 4. Testing uses a composition of 90% for training and 10% for testing. The test uses a variant of the GLCM, LBP extraction method, and a combination of both extraction methods, as shown in Table 1. The test results are shown in Figure 5.

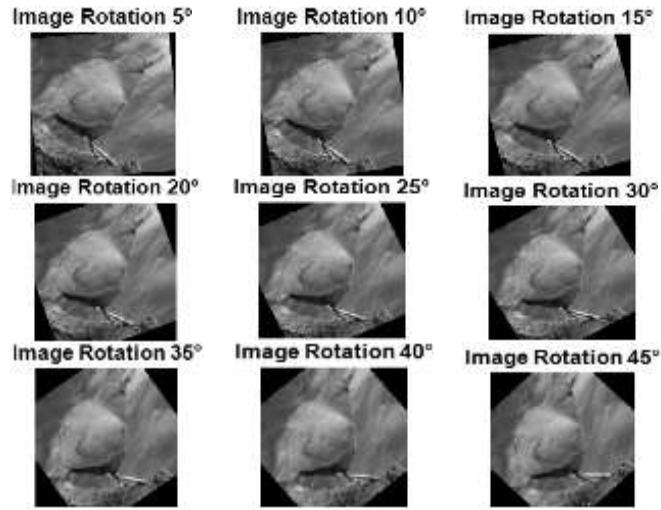

Figure 4. 2D object rotated 5 degrees

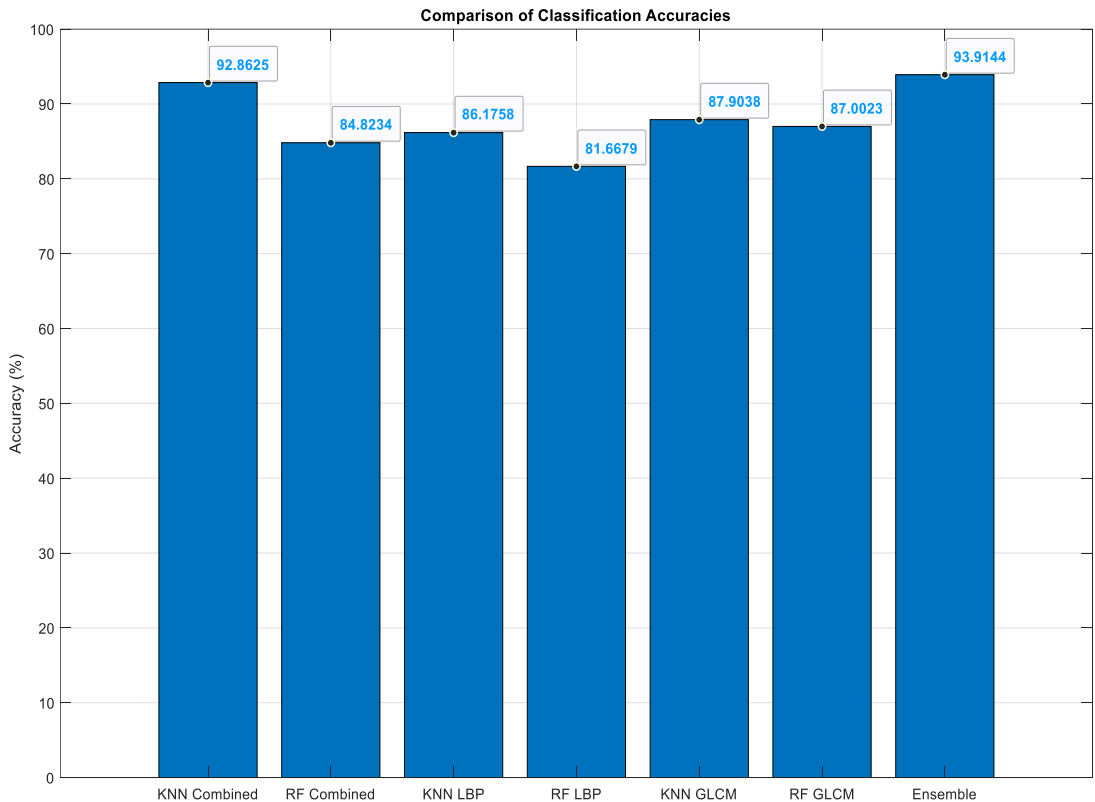

Figure 5. Variants of extraction and classification methods

The test uses several methods, namely GLCM, LBP, and a combination of GLCM and LBP. Likewise for classification using KNN, RF, and a combination which is represented as a VE. The test results are shown in Figure 5. For the combined extraction method with KNN classification, the accuracy value is 92.8625%, and the RF accuracy is 84.8234%. Meanwhile, for LBP extraction using KNN classification of 86.1758% and RF of 81.6679%. Meanwhile, the GLCM extraction method with KNN classification has an accuracy of 87.9038% and RF of 87.0023%. Meanwhile, for the extraction method combining GLCM with LBP and classification using VE, the accuracy reached 93.9144%. The overall results are shown in Figure 5. The confusion matrix shows that objects in each class can detect objects undergoing rotation and objects with variations in texture. The test results for accuracy, precision, recall, and F1-scores are presented in Table 2 and shown in Figure 6. Figure 6(a) KNN with LBP, Figure 6(b) RF with LBP, Figure 6(c) KNN with GLCM, Figure 6(d) RF with GLCM, Figure 6(e) KNN with combined, Figure 6(f) RF with combined, and Figure 6(g) VE.

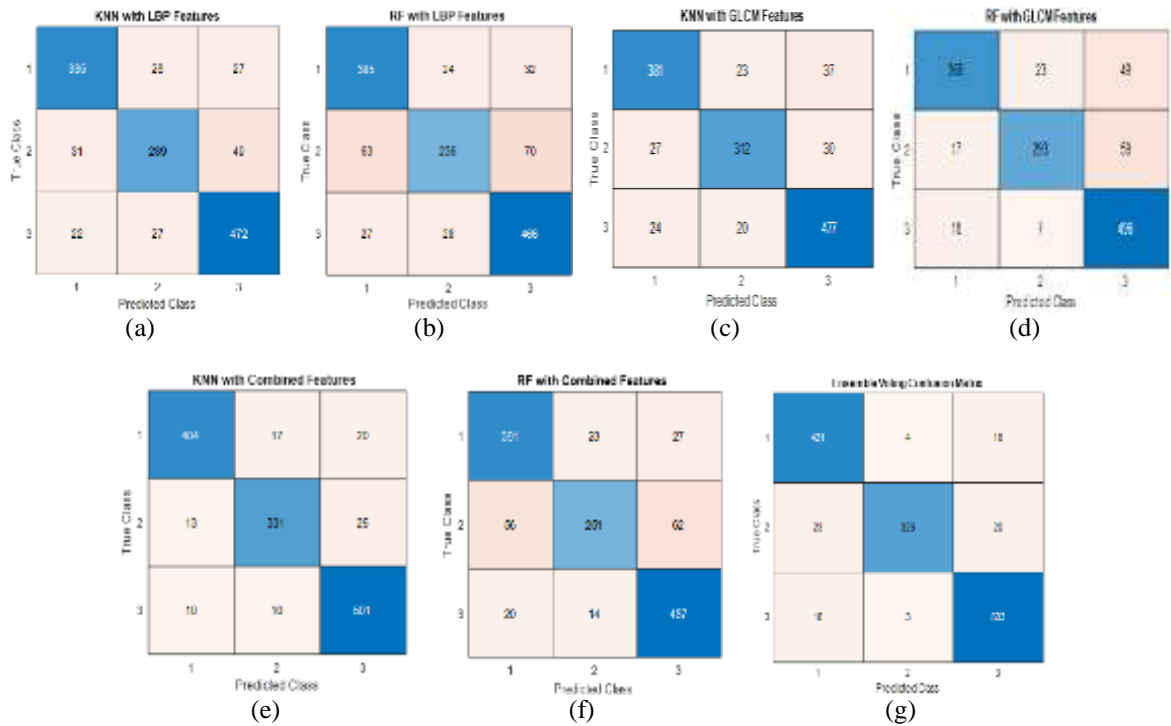

Figure 6. Confusion matrix 2D object detection model: (a) KNN with LBP, (b) RF with LBP,
(c) KNN with GLCM, (d) RF with GLCM, (e) KNN with combined, (f) RF with combined, and (g) VE

Table 2. Test results with variant extraction and classification

| Model | Accuracy (%) | Presisi (%) | Recall (%) | F1-score (%) |
| --- | --- | --- | --- | --- |
| Extraction combined and KNN | 92.65 | 92.68 | 92.65 | 92.64 |
| Extraction combined and RF | 84.82 | 85.00 | 84.82 | 84.47 |
| LBP and KNN | 86.18 | 86.14 | 86.18 | 86.11 |
| LBP and RF | 81.67 | 81.69 | 81.67 | 81.27 |
| KNN and GLCM | 87.90 | 87.91 | 87.90 | 87.88 |
| RF and GLCM | 87.00 | 87.56 | 87.00 | 86.93 |
| Extraction combined and VE | 93.91 | 94.06 | 93.91 | 93.90 |

The KNN classification method with combined features shows high results with 92.7% accuracy, 92.7% precision, 92.3% recall, and 92.5% F1-score. Demonstrates the model's ability to detect a minimal number of errors. Meanwhile, RF with combined features has lower accuracy than KNN, namely 84.8%, precision 85.1%, recall 83.4%, and F1-score 83.8%, indicating that RF is less effective than KNN in the detection model, with a tendency to make more errors in classification. Using the LBP feature in KNN has good performance with an accuracy of 86.2%, precision of 86.0%, recall of 85.5%, and F1-score of 85.7%, although there is a decrease in performance compared to using the combination feature. RF experienced a further decrease with the LBP feature, with an accuracy value of 81.7%, precision of 81.7%, recall of 80.2%, and F1-score of 80.5%.

The use of GLCM features provides better results than LBP for both classification methods. KNN with GLCM features achieved 87.9% accuracy, 87.9% precision, 87.5% recall, and 87.7% F1-score. RF showed improvement, with 87.0% accuracy, 88.1% precision, 86.1% recall, and 86.7% F1-score. Using the VE method with combination features obtained the highest accuracy, namely 93.9%, precision 94.3%, recall 93.5%, and F1-score 93.8%. This shows that the combination of predictions from each model (KNN and RF) can produce relatively high values. The use of combined features can improve the value better than the use of one type of feature, and the VE method is the most powerful approach in dealing with data variations and can effectively overcome the problem of rotation and texture variations in 2D images for detection.

## 4. CONCLUSION

The KNN method shows high effectiveness with an accuracy of 92.7% and an F1-score of 92.5%, reflecting a good balance between precision and recall. In contrast, RF is more sensitive to the type of features used; performance decreases with combined features, resulting in an accuracy of 84.8% and an F1-score of 83.8%. This indicates that feature variability influences the learning process and decisions taken by decision trees in RF. The LBP extraction method, which focuses on local textures, displays varying results. KNN with LBP features achieved 86.2% accuracy and 85.7% F1-score, while RF showed a decrease with 81.7% accuracy and 80.5% F1-score. This shows that KNN is more effective in classifying textures than LBP, while RF requires additional information or a different feature fusion strategy. When using the GLCM feature, KNN shows a small improvement compared to LBP, with an accuracy of 87.9% and an F1-score of 87.7%, while RF shows a more significant improvement with an accuracy of 87.0% and an F1-score of 86.7%. This indicates that GLCM features are more suitable for decision-based learning models such as RF. Finally, the VE classification model achieved the highest accuracy, 93.9%, and an F1-score of 93.8%, demonstrating its ability to reduce bias and increase generalization. This model was proven to be more robust and accurate compared to a single classification model, indicating that selecting appropriate features for the classification model can improve the overall performance of the 2D image detection system.

## BIOGRAPHIES OF AUTHORS

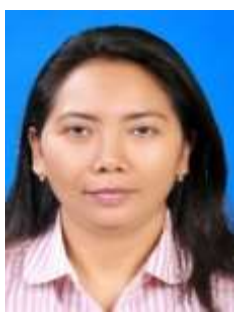

**Florentina Tatrin Kurniati** received her Master in Informatics Engineering from Atma Jaya Yogyakarta University (UAJY), Yogyakarta, Indonesia (2015), and is currently pursuing a doctoral program in computer science at Satya Wacana Christian University. Since 2008 he has been a lecturer and researcher at the faculty of informatics and computers, Institut Teknologi dan Bisnis STIKOM Bali, Indonesia. He is interested in adaptive noise cancellation, pattern recognition, object identification, and digital forensics. She can be contacted at email: 982022026@student.uksw.edu.

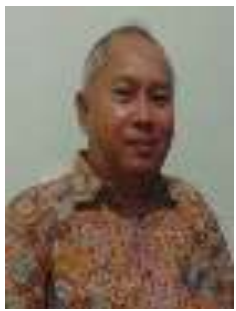

**Prof. Daniel HF Manongga** received an Engineer degree in electronics from Satya Wacana Christian University Salatiga in 1981, Master of Science in information technology from Queen Mary College-University of London in 1989, and Doctor of Philosophy in information systems, artificial intelligence and management science from Queen Mary College-University of London in 1996. His research areas include artificial intelligence, cloud computing, and the semantic web. Currently active as a teacher at the Faculty of Information Technology, Satya Wacana Christian University, Salatiga. He can be contacted at email: dmanongga@gmail.com.

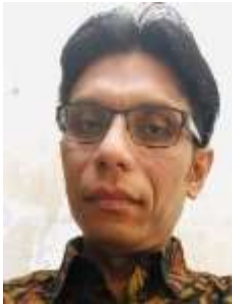

**Dr. Irwan Sembiring** completed his Bachelor's degree in 2001 at UPN "Veteran" Yogyakarta Indonesia. His Master degree is completed in 2004 from Gadjah Mada University Yogyakarta Indonesia, and Doctorate degree completed in 2016 at Gadjah Mada University Indonesia. Research interest in Network Security, Information System and Digital Forensic. Now he is a lecturer at faculty of Information Technology Satya Wacana Christian University, Salatiga, Indonesia. He can be contacted at email: irwan@uksw.edu.

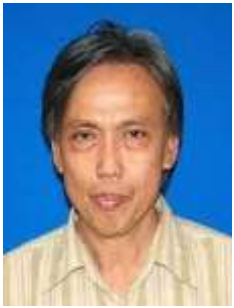

**Prof. Sutarto Wijono** completed his Bachelor's degree at Satya Wacana Christian University in 1987, his Master's degree at Universiti Kebangsaan Malaysia in 1997, and his Doctoral degree at the University of Indonesia in 1999. Currently active as a lecturer at the Faculty of Information Technology, Satya Christian University Wacana, Salatiga, specializing in Industrial and Organizational Psychology. He can be contacted at email: sutarto.wijono@uksw.edu.

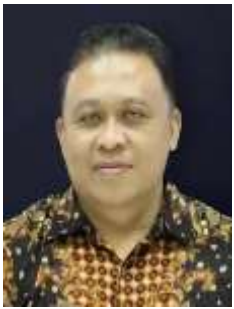

**Dr. Roy Rudolf Huizen, ST., MT**, Graduated with Doctor of Computer Science (2018) from Universitas Gadjah Mada (UGM) Yogyakarta, Indonesia. Lecturer and researcher at the Department of Magister Information System at the Institut Teknologi dan Bisnis STIKOM Bali, with research interests in the fields of object identification, signal processing, cyber security forensics and artificial intelligence. He can be contacted at email: roy@stikom-bali.ac.id.